\pdfoutput=1

\documentclass[11pt]{article}

\usepackage[]{emnlp2021}

\usepackage{times}
\usepackage{latexsym}

\usepackage[T1]{fontenc}

\usepackage[utf8]{inputenc}

\usepackage{microtype}
\usepackage[ruled]{algorithm2e}
\SetAlFnt{\small}

\usepackage[noend]{algpseudocode}
\usepackage{amsmath}
\usepackage{graphicx}
\usepackage{caption}
\usepackage{subcaption}
\usepackage{hyperref}
\usepackage{empheq}
\title{DRIFT: A Toolkit for Diachronic Analysis of Scientific Literature}

\author{Abheesht Sharma\thanks{\scriptsize\hspace{2mm}Equal contribution. Author ordering determined by coin flip.} \\
Dept. of CS\&IS \\
BITS Pilani, Goa Campus \\
 \texttt{\scriptsize f20171014@goa.bits-pilani.ac.in}
 
\And

Gunjan Chhablani\footnotemark[1] \\
Dept. of CS\&IS \\
BITS Pilani, Goa Campus \\
 \texttt{\scriptsize chhablani.gunjan@gmail.com}

\AND

Harshit Pandey\footnotemark[1] \\
    Dept. of Computer Science \\
    Pune University \\
    \texttt{\scriptsize hp2pandey1@gmail.com} 

\And
 Rajaswa Patil \\
  Dept. of E \& E Engineering\\
  BITS Pilani, Goa Campus\\
  \texttt{\scriptsize f20170334@goa.bits-pilani.ac.in}
}

\begin{document}
\maketitle
\begin{abstract}
In this work, we present to the NLP community, and to the wider research community as a whole, an application for the diachronic analysis of research corpora. We open source an easy-to-use tool coined \emph{DRIFT}, which allows researchers to track research trends and development over the years. The analysis methods are collated from well-cited research works, with a few of our own methods added for good measure. Succinctly put, some of the analysis methods are: keyword extraction, word clouds, predicting declining/stagnant/growing trends using Productivity, tracking bi-grams using Acceleration plots, finding the Semantic Drift of words, tracking trends using similarity, etc. To demonstrate the utility and efficacy of our tool, we perform a case study on the \emph{cs.CL} corpus of the arXiv repository and draw inferences from the analysis methods. 
The toolkit and the associated code are available \href{https://github.com/rajaswa/DRIFT}{here}.
\end{abstract}

\section{Introduction}
\label{sec:introduction}
Historians  perform comparative studies between the past and the present to explain certain phenomena. Studying the past also helps us in making plausible predictions about the future.
Major sources of information about the past are old-age texts, tomes and manuscripts. Language changes and shifts with changes in society and culture. For example, words such as \emph{curglaff} and \emph{lunting} have become obsolete, the word \emph{gay}'s meaning has changed entirely and new words such as \emph{LOL} and \emph{ROFL} have gained prominence with the emergence of the ``texting generation''.

In the research world, analysis of old articles and papers is gaining importance. Analysing old documents can revive research topics which have been forgotten over the decades; removing the cobwebs on these topics can inspire path-breaking research. Explanations of why we arrived at certain conclusions can also be gleaned by peeping into the past. Performing such diachronic analysis of text has become easy because of the rapid growth of NLP. Temporal word embeddings such as TWEC~\cite{dicarlo-2019-twec} enable us to do so.

In this paper, we introduce a hassle-free application for the diachronic analysis of research corpora. We name our application \emph{DRIFT}, an acronym for \textbf{D}iach\textbf{R}onic Analysis of Scient\textbf{IF}ic Li\textbf{T}erature. \emph{DRIFT} provides researchers a one-click way to perform various diachronic analyses.

Our contribution is two-fold: 1) We assemble (and propose) different methods for diachronic analysis of research corpora; 2) We make it extremely convenient for researchers to analyse research trends; this, in turn, will encourage researchers to find latent, quiescent topics which may have huge research potential in the near future. The main principles behind the design for \emph{DRIFT} are \textbf{ease of usage} and \textbf{flexibility}.

\begin{figure*}
    \centering
    \includegraphics[scale=0.97]{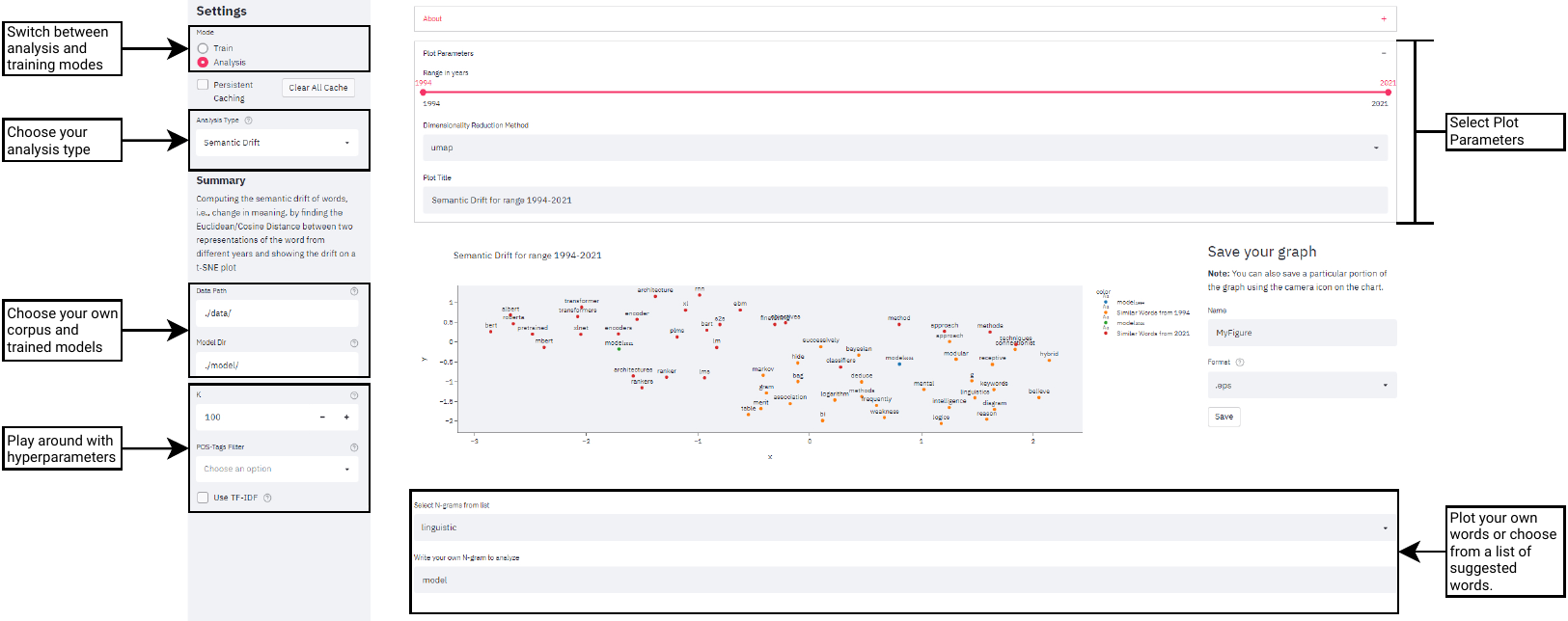}
    \caption{Snapshot for the application dashboard.}
    \label{fig:dashboard}
\end{figure*}

The rest of the paper is organised as follows.
In Section~\ref{sec:related_works}, we perform an extensive literature survey.
Section~\ref{sec:datasets} describes the research corpora and the methods for crawling the corpora.
In Section~\ref{sec:methodology}, we explain the training methodology for obtaining diachronic embeddings and our easy-to-use application's dashboard and layout in detail.
Section~\ref{sec:analysis} elucidates the various analysis methods which researchers can employ with a mere click of a button.

\section{Related Works}
\label{sec:related_works}
\subsection{Diachronic Embeddings}
Savants have attempted to leverage word embeddings to study semantic changes of words over time~\cite{hamilton-etal-2016-diachronic,kulkarni-2014-statistically}. The major issue has been the inability to compare word vectors of two different time periods since word embedding algorithms are stochastic in nature and invariant under rotation, i.e., they belong to different coordinate systems~\cite{kutuzov-etal-2018-diachronic}. 
Therefore, attempts have been made to align these vectors~\cite{szymanski-2017-temporal,schlechtweg-etal-2019-wind,bamman-etal-2014-distributed}. 
The most recent attempt to resolve this issue is TWEC~\cite{dicarlo-2019-twec}, which implicitly aligns the word vectors. We use TWEC in our work.

\subsection{Analysis Attempts and Tools for Research Corpora}
Many researchers have devoted their time to analysing different research corpora with respect to time. Most of the analysis is citation-based, or the researchers analyse the diversity of the research world (age-based, gender-based analysis)~\cite{mohammad-2019-state}.
In some, authors have introduced new research corpora such as the NLP4NLP Corpus~\cite{mariani-2019-nlp4nlp}. 
\citet{gollapalli-li-2015-emnlp} perform comparative analyses between two conferences such as EMNLP and ACL by expressing both venues as a probability distribution over time. 
A good amount of research has also focused on statistical techniques. For example, \citet{schumann-2016-brave} uses frequency and productivity to identify emerging/stagnant/falling topics. 
Others have devised Machine Learning/Deep Learning methods to identify future trends~\cite{francopoulo-etal-2016-predictive}.

NLP Scholar~\cite{mohammad-2020-nlp-scholar} is an interactive, visual tool which makes it simpler for researchers to find impactful scientific articles in bygone years from the ACL Anthology based on the number of citations, searching for relevant related works, study the changes in the number of articles and citations, etc.

While significant effort has gone into procuring research corpora and in devising different analysis methods, not much work has been done in building a user-friendly, convenient application which ties up statistical, learning-based analysis and temporal embedding-based methods and makes it easy for the research community to draw inferences and identify research trends.

\section{Datasets}
\label{sec:datasets}
arXiv\footnote{\url{https://arxiv.org/}} is a free-access repository of e-prints of scientific papers, articles, essays and studies, launched in 1991 by Cornell University. 
arXiv covers a vast array of disciplines such as Computer Science, Mathematics, Physics, Chemistry, etc. 
We perform our experiments and analysis on a burgeoning sub-domain of Computer Science, namely, Computation and Language (\emph{cs.CL}).

Moreover, we perform our analyses on abstracts. The rationale behind this is: 1) The abstract of a paper condenses and encapsulates the whole paper and conveys its major contributions; 2) The rest of the paper may have tables, figures, arcane mathematical equations, intricacies and esoteric jargon which are difficult to work with.

We restrict our analysis to the 1994-2021 period for the following reasons: 1) arXiv has cs.CL and cs.CV papers from 1994 onwards; 2) We considered crawling the abstracts of papers published prior to 1994 from their PDFs (we obtained the URLs from the ACL Anthology\footnote{{\scriptsize \url{https://github.com/acl-org/acl-anthology}}}). However, this proved to be no mean feat, considering that the old PDFs used different font styles, unselectable texts, etc. The procured abstracts had grammatical flaws.

We use the arXiv API to obtain the metadata of the papers (specifically, URL, date of submission, title, authors, and abstract). To ensure that low quality articles are not included, we sort the papers by their relevance. In total, we analyse 27,384 abstracts.

\section{Methodology}
\label{sec:methodology}
\subsection{Diachronic Embeddings}
\label{ssec:diachronic_embeddings}
As mentioned earlier, we use TWEC~\cite{dicarlo-2019-twec} (Temporal Word Embeddings with Compass) to create dynamic word embeddings for each time slice. TWEC uses Word2Vec~\cite{mikolov2013efficient} and trains it on data for all the years to learn atemporal embeddings. Then, the ``pre-trained'' target embeddings are frozen, which serves as the ``compass'' and temporal embeddings are learned for each year as context embeddings. We use this method because it scales well with large corpora, and it also makes the training simple and efficient.

\subsection{Application}
\label{ssec:dashboard}
We build our application using Streamlit\footnote{\url{https://streamlit.io/}}, a framework in Python for making data applications. An overview of the application dashboard is shown in Figure~\ref{fig:dashboard}. The dashboard has two modes: \emph{Train} and \emph{Analysis}. The general workflow of the application is shown in Figure~\ref{fig:system_design} and described as follows:

\begin{figure}[htbp]
    \centering
    \includegraphics[scale=0.5]{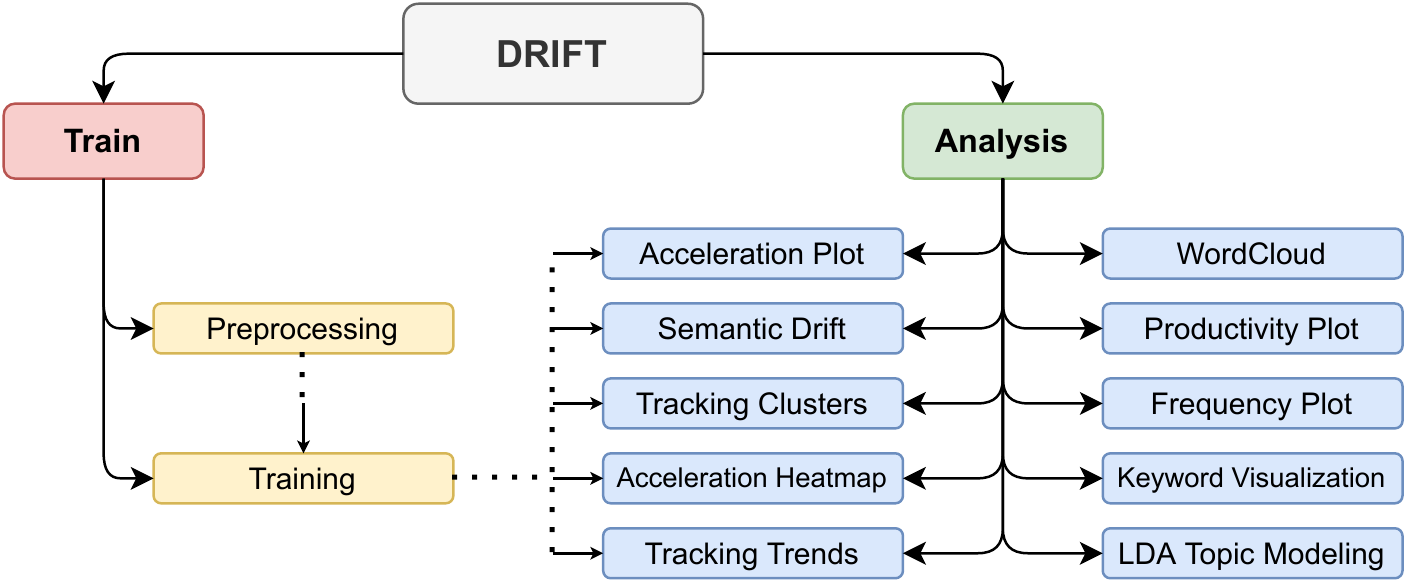}
    \caption{System design for the application.}
    \label{fig:system_design}
\end{figure}

\subsubsection{\emph{Train} Mode}
For flexibility, in the \emph{Train} mode, we give users an option to upload their corpus and train the TWEC model with a click of a button. The sidebar has two subsections.

\paragraph{Preprocessing} Before training TWEC, the dataset has to be preprocessed. The preprocessing pipeline we employ comprises simple steps: convert to lower-case, lemmatisation, remove punctuation and stopwords, remove non-alphanumerics, etc. We also remove words which frequently appear in research papers such as \emph{paper}, \emph{systems}, \emph{result}, \emph{approach}, etc. This subsection has three input text boxes: \emph{JSON Path} for the input dataset/corpus, \emph{Text Key} for the key whose corresponding value contains raw text, \emph{Data Path}, where the preprocessed data is to be stored. On clicking the button \emph{Preprocess}, the raw text is preprocessed.

\paragraph{Training} TWEC hyperparameters such as embedding size, number of static iterations, dynamic iterations, negative samples, window size are presented as text boxes/drop-down menus here. On clicking the button \emph{Train}, the TWEC model is trained on the corpus.
Options/hyperparameters/input paths for preprocessing and TWEC training are presented as text boxes/drop-down menus in the sidebar. 

\subsubsection{\emph{Analysis} Mode}
The \emph{Analysis} mode is the most vital component of the application. The analysis method can be chosen from a drop-down menu. Once an analysis method is chosen, the sidebar displays several parameters which generally include data path, K (number of keywords to choose from compass), whether to use TF-IDF or use particular Parts-of-Speech, etc. As an example, method-specific parameters for \textit{Semantic Drift} include: model path,  K(sim.) (top-k most similar words to be shown around chosen words), K(drift) (allows user to select from top-k most drifted words), and distance metric (euclidean/cosine distance).

The page for every analysis method contains an expander, i.e., a collapsible section which has options such as a slider to decide the range of years to be considered for analysis, or a text box for a particular word to be analysed. Below the expander, interactive graphs, t-SNE~\cite{Maaten2008VisualizingDU}/UMAP~\cite{mcinnes2020umap}/PCA plots and tables are displayed, based upon user's selection. Every button, text box, drop-down menu has a ``help'' option to guide users on its usage and purpose. 

On the right-hand side, there is an option to export graphs in different formats such as PDF, SVG, etc. for easy integration with research papers.

\section{Analysis Methods and Discussion}
\label{sec:analysis}
\begin{figure*}
    \centering
    \includegraphics[width=\textwidth]{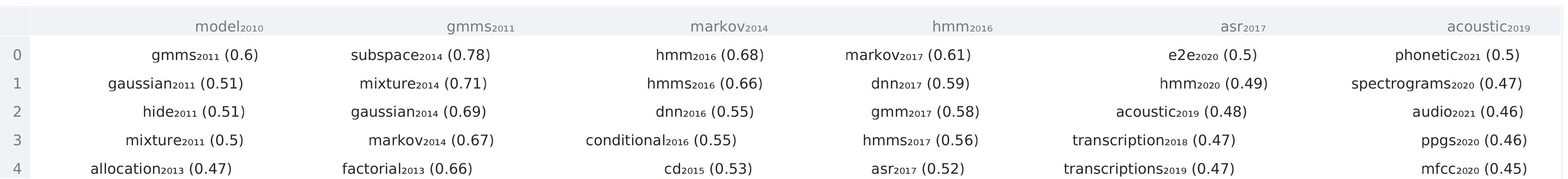}
    \caption{Tracking the word \textit{``model''} from the year 2010 to the year 2019 with stride=3.}
    \label{fig:track_trends_with_sim}
\end{figure*}

\subsection{Word Cloud}

A Word Cloud is a graphical representation of the keywords of a corpus, i.e., words which have higher frequency are given more importance. This importance is translated in terms of size and colour in the visualisation.
We give the user an option to choose the year for which the analysis is to be done. Other options in the sidebar include minimum and maximum font size, number of words to be displayed, colour, width and height of the word cloud. The user can select the year from the expander. On the main page, we display the word cloud. An example is shown in Figure~\ref{fig:word_cloud}.

\subsection{Productivity/Frequency Plot}

\citet{schumann-2016-brave} uses normalized term frequency 
and term productivity as measures for identifying growing/consolidated/declining terms.
Term productivity is a measure of the ability of a concept to produce new multi-word terms. 
In our case, we use bigrams. 
For each year $y$ and single-word term $t$, and associated $n$ multi-word terms $m$, the productivity is given by the entropy:
\begin{equation}
    e(t,y) = - \sum_{i=1}^{n} \log_{2}(p_{m_{i},y}).p_{m_{i},y}
\end{equation}
where
\begin{equation}
    p_{m,y} = \frac{f(m)}{\sum_{i=1}^{n}f(m_{i})}
\end{equation}
and $f(m)$ is frequency of the term.
Based on these two measures (over the years), words are clustered into three categories:

\begin{itemize}
    \item \textbf{Growing Terms}: Those which have increasing frequency and productivity in the recent years.
    \item \textbf{Consolidated Terms}: Those that are growing in frequency, but not in productivity.
    \item \textbf{Terms in Decline}: Those which have reached an upper bound of productivity and have low frequency.
\end{itemize}

In the sidebar, the user can choose the $N$ in N-gram and the $K$ in Top-K. The default for keyword extraction in all the methods is normalised frequency, but the user can opt for frequency or TF-IDF. Alternatively, the user can also choose the POS tag for filtering the keywords.

In the expander, the user can select words from the suggested keywords or add their own words for analysis. In the main section, the productivity and normalised frequency graphs are displayed. Below these graphs, we have a dataframe which displays which clusters the words belong to. An example is shown in Figure~\ref{fig:productivity}.

\subsection{Acceleration Plot}
\label{sub_sec:acc_plot}
Inspired by \citet{dridi-2019-deephist}, we analyse the semantic shift of a pair of words with respect to each other. We aim to identify fast converging keywords using ``acceleration''. The acceleration matrix is calculated as the difference between similarity matrices of two different time periods, where each entry is:
\begin{multline}
    acc(w_i, w_j)^{t\rightarrow(t+1)} = \\
    sim(w_i, w_j)^{(t+1)} - sim(w_i, w_j)^{t}
\end{multline}

where $(w_i, w_j)$ is the word pair being analysed, $sim(w_i, w_j)$ is the cosine similarity between words $w_i$ and $w_j$, and $(t, t+1)$ is the time range.
If the words $w_i$ and $w_j$ converge, the cosine similarity value between them will increase, or in other words, $acc(w_i, w_j)^{t\rightarrow(t+1)} > 0$.
We pick the word pairs with the highest acceleration values. 

\begin{figure*}[t!]
    \centering
    \includegraphics[scale=1]{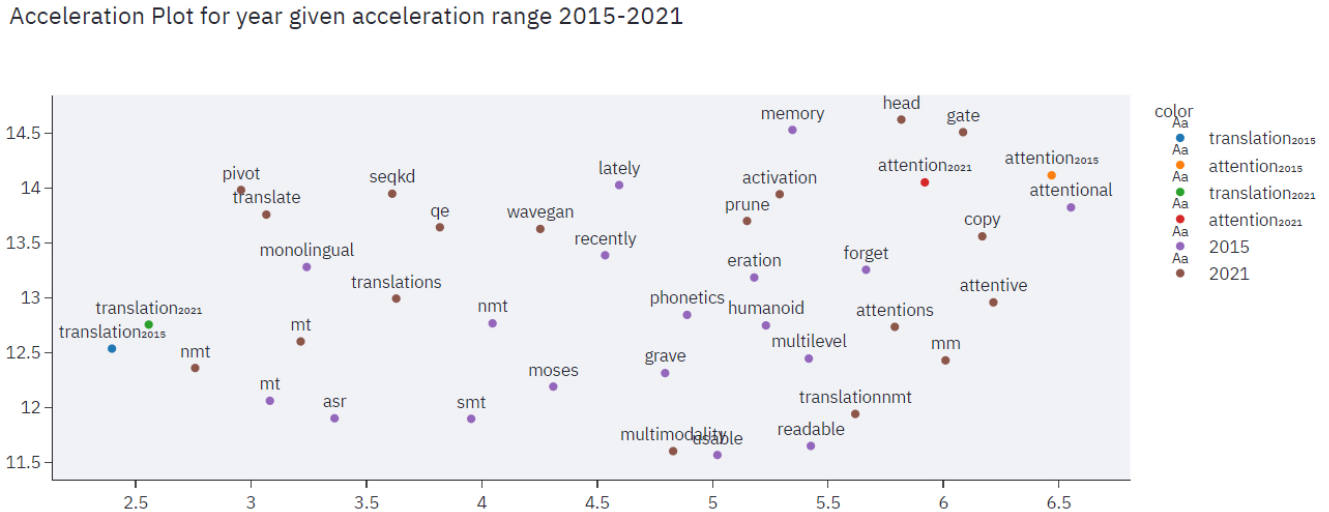}
    \caption{Convergence of the words: \textit{``Translation''} and \textit{``Attention''} (2015-2021).}
    \label{fig:acc_plot_translation_linguistic}
\end{figure*}

To demonstrate the convergence of a pair of words, we plot the pair on a two-dimensional plot, along with the top-K most similar words around them. An example is shown in Figure~\ref{fig:acc_plot_translation_linguistic}.

In addition to $K$, POS Tag Filters and an option to opt for TF-IDF, we also have $K(acc.)$ and $K(sim.)$. The parameter, $K(sim.)$, is for deciding how many words to plot around the two chosen words. $K(acc.)$ is for finding the top-K most accelerated pair of words.

The expander has a slider for selecting the range of years, below which there is a dataframe, listing the top-K most accelerated keywords. There are two drop-downs for selecting the pair of words to be analysed.

The main section has the graph of the embedding space, showing the convergence/divergence of the chosen pair of words. 

\subsection{Semantic Drift}

The semantics of words perpetually change over time, i.e., words drift from one point to another. 
This incessant \emph{semantic drift} is a product of changing social and cultural norms, transition in linguistic usage, amongst other factors.
Since we use TWEC embeddings, the representations of a word across time periods are implicitly aligned and we can directly compute the shift as the distance between the two representations.

We compute the drift as the Euclidean Distance or the Cosine Distance between the first year embedding of the word and the last year embedding of the word. We sort the words according to their drift (in descending order). For more details on the algorithm, refer to Subsection~\ref{appssec:sem_shift} in the Appendix.

In order to visualise the drift, we plot the two representations of the word on a UMAP/tSNE/PCA plot, along with the most similar words in the respective years which are clumped around the two representations. Plotting the top-k most similar words helps the user in analysing the shift in meaning.

Other than the usual parameters - $K$, POS Tag Filters, TF-IDF, the sidebar has options to choose $K(sim.)$ and $K(drift)$. The expander has a slider for selecting the range of years.
Below the graphs, a drop-down menu has the top-K(drift) most drifted words as suggestions. The user can also input a custom word.

\begin{figure*}
    \centering
    \includegraphics{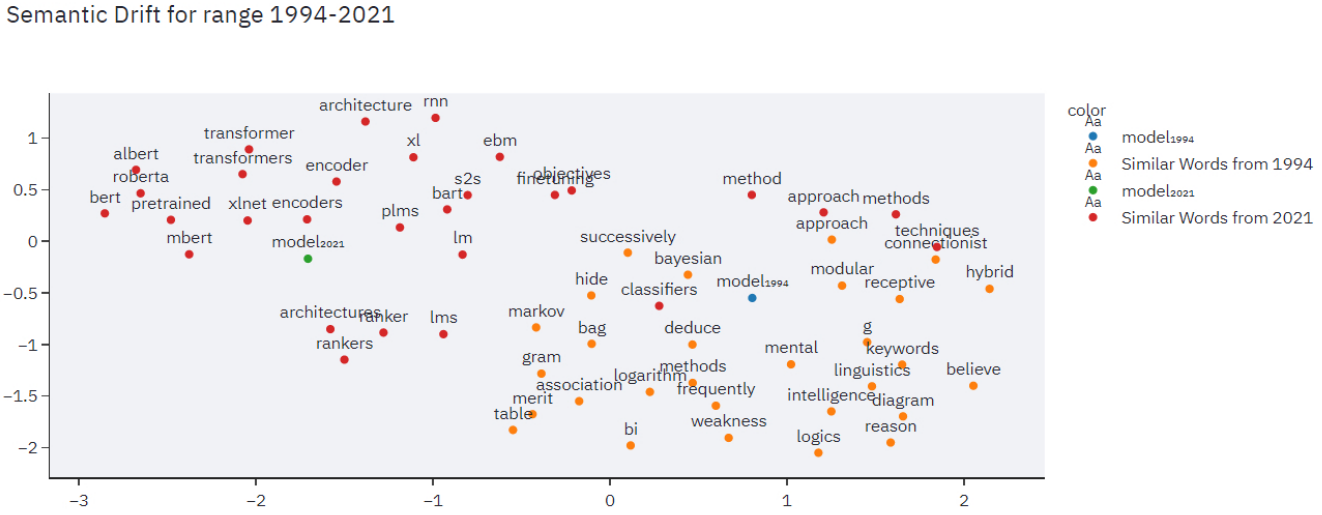}
    \caption{Semantic Drift of the word \textit{``model''} (1994-2021).}
    \label{fig:semantic_drift_model}
\end{figure*}

A couple of examples are shown in Figure~\ref{fig:semantic_drift_model}. Clearly, the word ``model'' has drifted from the vicinity of \emph{bayesian}, \emph{markov}, \emph{bag}, \emph{gram}, \emph{logics} in 2017 to \emph{albert}, \emph{roberta}, \emph{xlnet}, \emph{mbert}, \emph{bart} in 2021. This makes sense, because back in 1994, a conventional NLP model was related to bag-of-words, and Markov/HMM-based~\cite{fosler-1998-markov}/Bayesian methods, while in 2021, the focus has shifted to RoBERTa~\cite{liu-2020-roberta}, XLNet~\cite{zhilin-2019-xlnet}, BART~\cite{lewis-etal-2020-bart} and other transformer models.

\subsection{Tracking Clusters}

Word meanings change over time. They come closer or drift apart. In any given year, certain words are clumped together, i.e., they belong to one cluster. But over time, clusters can break into two/coalesce together to form one. Unlike the previous module which tracks movement of one word at a time, here, we track the movement of clusters.

Our implementation is simple. Given a range of years, and some common keywords, we cluster the words for every year, and display the UMAP/tSNE/PCA graph (with clusters) for every year separately. We use the KMeans Clustering Algorithm. A visualisation is shown in Figure~\ref{fig:track_clusters}.

The additional options in the sidebar include number of clusters and max. number of clusters. If number of clusters is set to 0, the application finds the optimal number by searching between $[3,Max.\;Number\;of\;Clusters]$ using Silhouette score. Additionally, we have an option to choose the library of implementation: faiss\footnote{{\scriptsize\url{https://github.com/facebookresearch/faiss}}}/scikit-learn\footnote{{\scriptsize\url{https://github.com/scikit-learn/scikit-learn}}}. Faiss is up to 10 times faster than sklearn.

\subsection{Acceleration Heatmap}
We use the same \textit{acceleration} measure as in Section~\ref{sub_sec:acc_plot}. Instead of listing the top-K pairs based on acceleration, we plot a heatmap of the acceleration between two years. This can be useful in analysing how different word-pairs are converging at different rates between two years. This heatmap can be used in combination with the Acceleration Plot, where a representation of the word-pairs selected based on the heatmap can be explored. Refer to Figure~\ref{fig:acc_heatmap} to get an idea of the visualisation.

\subsection{Track Trends with Similarity}

We wish to chart the trajectory of a word/topic from \emph{year 1} to \emph{year 2}. 
To accomplish this, we allow the user to pick a word from \emph{year 1}. At the same time, we ask the user to provide the desired \emph{stride}. We search for the most similar word in the next $stride$ years. We keep doing this iteratively till we reach \emph{year 2}, updating the word at each step via user input from the top-K similar words. An overview of the algorithm is given in Subsection~\ref{appssec:track_trends}. An example is given in Figure~\ref{fig:track_trends_with_sim}. The trajectory we follow is model, gmms, markov, hmm, asr, phonetic. This makes sense, because HMMs, GMMs were used for ASR in the past.

 

\subsection{Keyword Visualisation}
Here, we use the YAKE Keyword Extraction method~\cite{campos-2020-yake} to extract keywords. Yake Score is indirectly proportional to the keyword importance. Hence, we report the scores differently\footnote{$new\_score = \frac{1}{10^{5} \times yake\_score}$}.

The user can select $Max.\;N-gram$ in the sidebar. The year can be selected from the slider in the expander. The main section has a bar graph, with n-grams on the y-axis and their importance scores on the x-axis. A visualisation is shown in Figure~\ref{fig:keyword_viz}.

\subsection{LDA Topic Modelling}

Latent Dirichlet Allocation (LDA) \cite{blei-2003-latent} is a generative probabilistic model for an assortment of documents, generally used for topic modelling and extraction. 
LDA clusters the text data into imaginary topics. 
Every topic can be represented as a probability distribution over n-grams and every document can be represented as a probability distribution over these generated topics. 

We train LDA on a corpus where each document contains the abstracts of a particular year. We express every year as a probability distribution of topics. 
An example is shown in Figure~\ref{fig:lda_fig}.

\section {Accessibility}
\label{sec:accessibility}
The code for the toolkit is open-sourced, and is distributed under a MIT License. The latest stable code release is available \href{https://github.com/rajaswa/DRIFT/releases/tag/v0.1.0}{here}. An online demo of the application is hosted \href{https://share.streamlit.io/rajaswa/drift/main/app.py}{here}. A set of short video demonstrations can be found \href{https://github.com/rajaswa/DRIFT/blob/main/README.md}{here}.

\section {Conclusion and Future Work}
\label{sec:conclusion}
In this paper, we present \textit{DRIFT}, a hassle-free, user-friendly application for diachronic analysis of scientific literature. We compile well-cited research works and provide an interactive and intuitive user-interface using Streamlit. We perform a case-study on \textit{cs.CL} corpus from arXiv repository, and demonstrate the effectiveness and utility of our application in analysing such corpora. As an extension of our work, apart from adding upcoming analysis methods (citation-based/statistical/knowledge graphs), we also intend to make our application modular to make it easier for users to add their own methods, provide semi-automated inference from the graphs/tables, and allow easy integration with \LaTeX.

\bibliography{anthology,custom}
\bibliographystyle{acl_natbib}

\appendix

\section{Algorithms}
\label{appsec:algorithms}

\subsection{Semantic Shift}
\label{appssec:sem_shift}

An overview of the approach is given in Algorithm~\ref{alg:semantic_shift}. The $dist(x, y)$ function can either be the Euclidean Distance or Cosine Distance. In the algorithm, we find the most drifted words from a given list of words by sorting them according to the distance. $year\_x\_model$ is the aligned TWEC Model for year $x$.

\begin{algorithm}
\caption{Semantic Shift}
        \label{alg:semantic_shift}
        \KwData{$words, year\_1, year\_2$}
		\KwResult{$word\_dist$}
		\BlankLine
		$word\_dist \gets \{\}$
		\BlankLine
        \For{$word\;in\;words$}
        {
            $year\_1\_emb \gets year\_1\_model(word)$
    
            $year\_2\_emb \gets year\_2\_model(word)$
            
            $word\_dist[word] \gets dist(year\_1\_emb, year\_2\_emb)$
        }
        \BlankLine
        $word\_dist \gets sort\_by\_value(word\_dist, ``desc")$
		
\end{algorithm}

\subsection{Track Trends with Similarity}
\label{appssec:track_trends}

\begin{algorithm}
\caption{Track Trends with Similarity}
        \label{alg:tracking_trends_with_similarity}
        \KwData{$word, year\_1, year\_2$}
		\KwResult{$word\_traj$}
		\BlankLine
 
         $year\gets year\_1$\\
        $word\_iter \gets word$\\
        $word\_traj \gets [(word, year\_1)]$\\
        \While{$year\leq year\_2$}
        {
            $word\_iter \gets MSW(word\_iter,range=[year+1, year+stride])$\\
            $year \gets get\_year(word\_iter)$\\
            $word\_traj.append((word\_iter, year))$
        }
\end{algorithm}

Algorithm~\ref{alg:tracking_trends_with_similarity} gives an overview of our approach. $MSW(word, range=[year+1, year+stride])$ finds the most similar word to $word$ in the years ${year+1, year+2, ..., year+stride}$. Here, we have demonstrated (for the sake of simplicity) that the most similar word is chosen at every step. In the actual implementation, we allow the user to choose from the top-K most similar words at every step. $stride$ is available as an argument in the sidebar, along with $K(sim.)$. In the main section, a table (which is dynamically updated) displays the trajectory taken by the initial word.


\section{Additional Demonstrative Analyses}
\label{appsec:additional_egs}

\begin{figure}[htbp]
     \centering
     \includegraphics[width=\columnwidth]{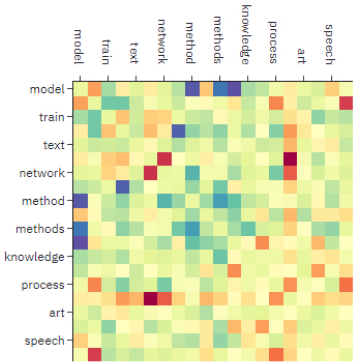}
     \caption{Acceleration Heatmap for top-K words between two years. A darker red colour implies a higher positive acceleration value.}
     \label{fig:acc_heatmap}
\end{figure}

\begin{figure*}[ht]
  \begin{subfigure}[t]{\columnwidth}
    \centering
    \includegraphics[width=\columnwidth]{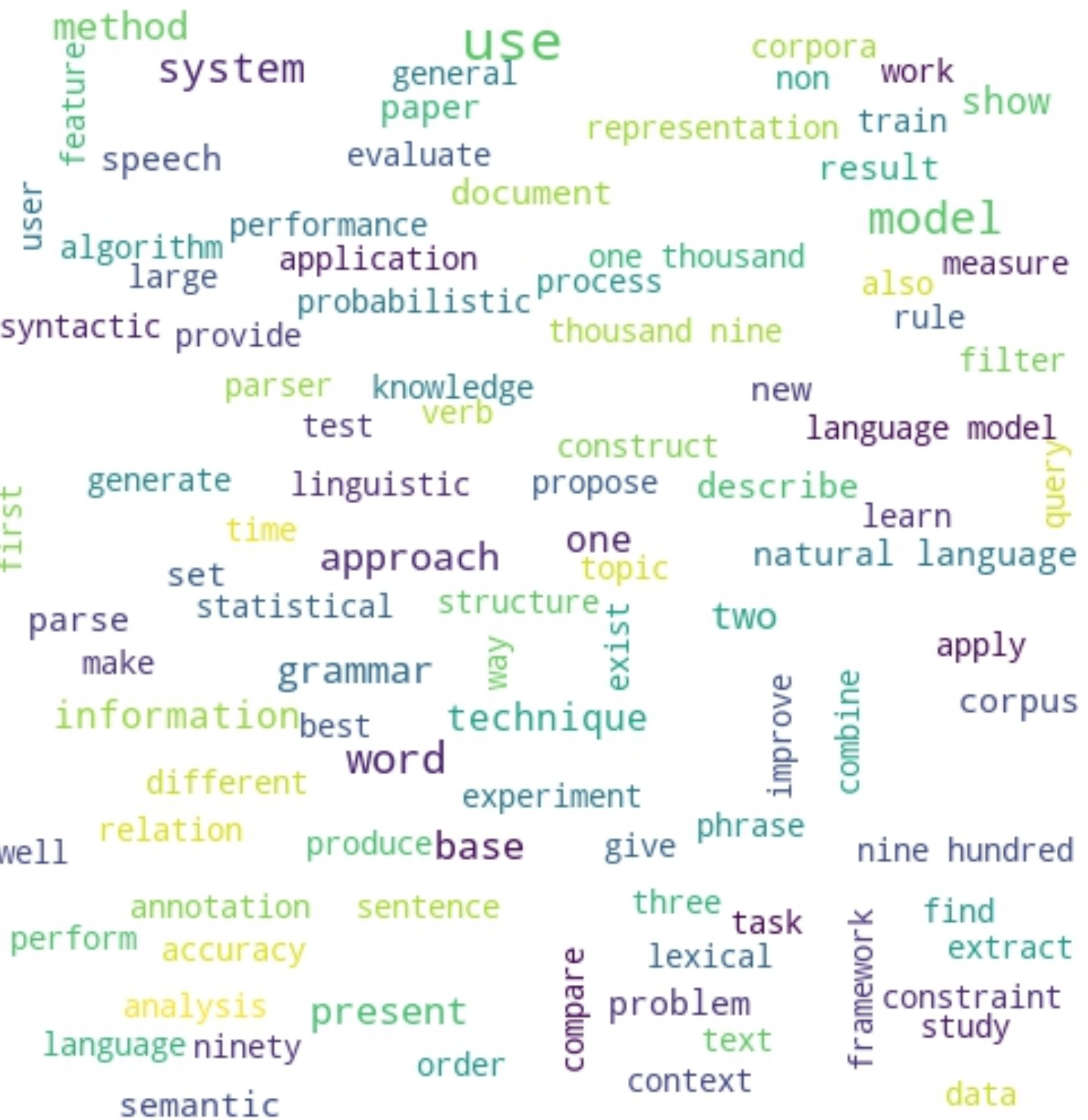}
    \caption{Word Cloud for the year 2000.}
    \label{fig-wc-2000}
  \end{subfigure}
  \hfill
  \begin{subfigure}[t]{\columnwidth}
    \centering
    \includegraphics[width=\columnwidth]{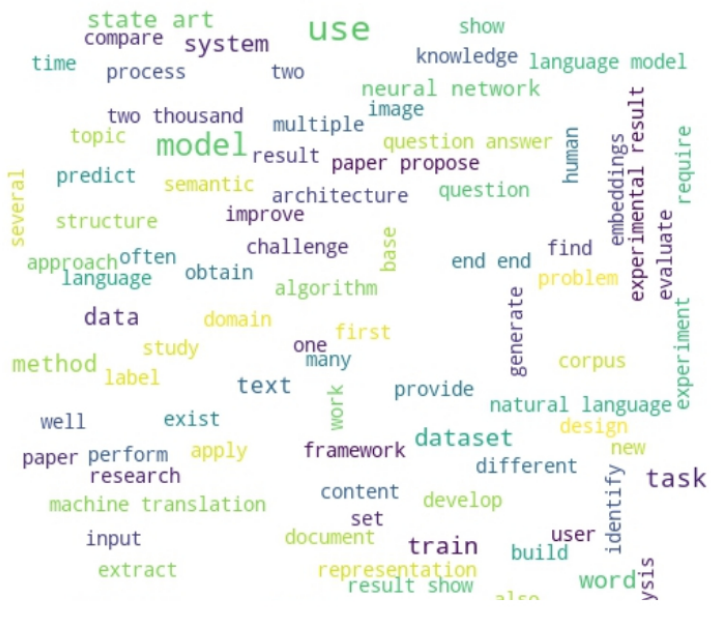}
    \caption{Word Cloud for the year 2018.}
    \label{fig-wc-2018}
  \end{subfigure}
  \caption{Word Cloud Demonstrations.}
  \label{fig:word_cloud}
\end{figure*}

\begin{figure*}[htbp]
     \centering
     \includegraphics[width=\textwidth]{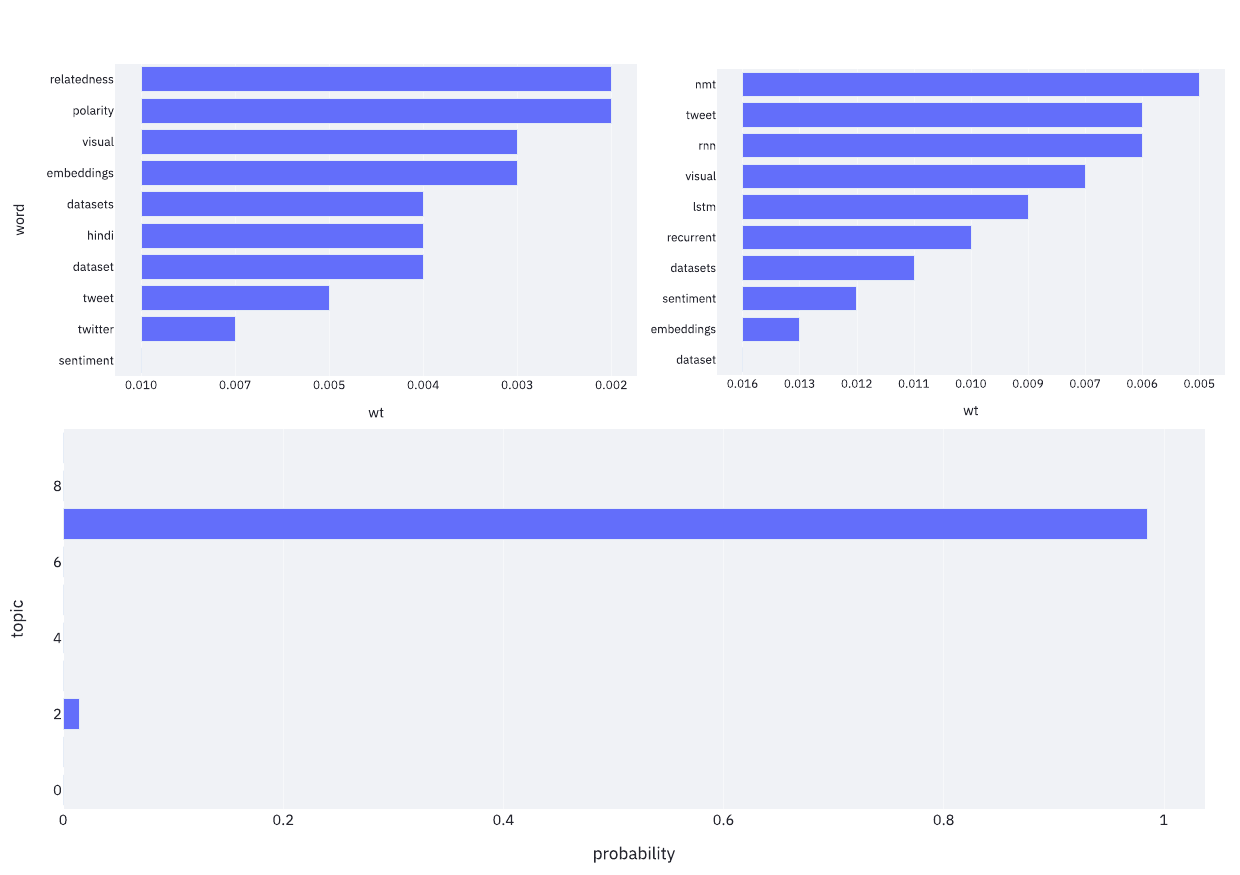}
     \caption{LDA Topic Modeling for the year 2014. The major topics in 2014 are 2 and 7. Topic 2 is primarily composed of ``visual'', ``embeddings'', etc. and Topic 7 is composed of ``nmt'', ``tweet'', ``rnn'', etc.}
     \label{fig:lda_fig}
\end{figure*}


\begin{figure*}[htbp]
    \centering
    \includegraphics[width=\textwidth]{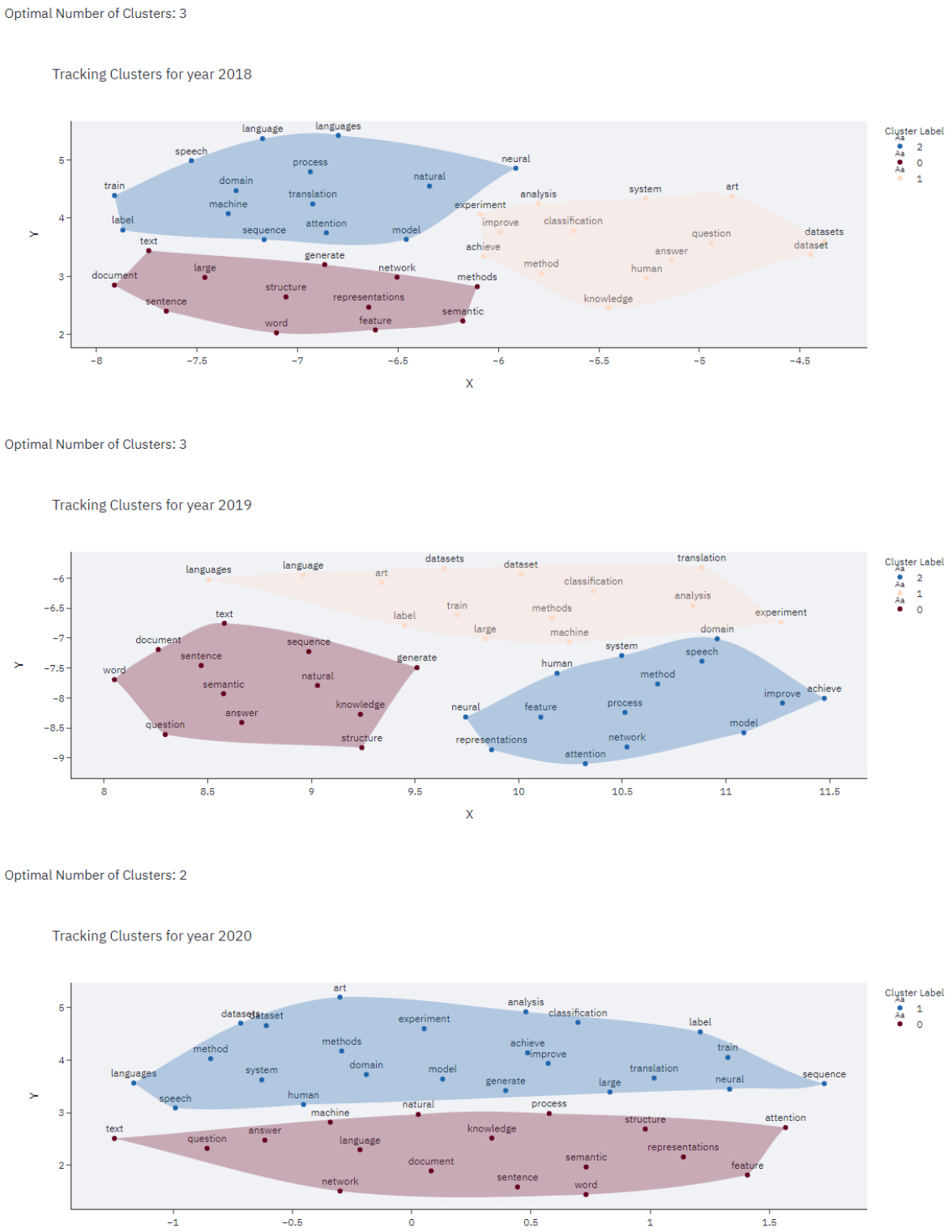}
    \caption{Tracking Clusters from 2018 to 2020.}
    \label{fig:track_clusters}
\end{figure*}

\begin{figure*}[thbp]
    \centering
    \includegraphics[width=\textwidth]{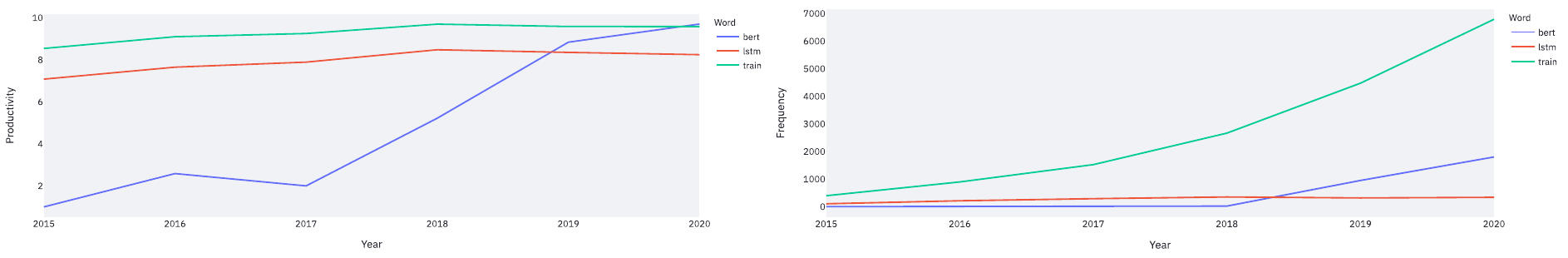}
    \caption{Productivity and Normalised Frequency Plots for the words ``BERT'', ``LSTM'', ``train'' (2015-2020). We can classify ``BERT'' as a growing term, ``train'' as a consolidated term and ``LSTM'', a  declining term.}
    \label{fig:productivity}
\end{figure*}

\begin{figure*}[thbp]
    \centering
    \includegraphics[width=\textwidth]{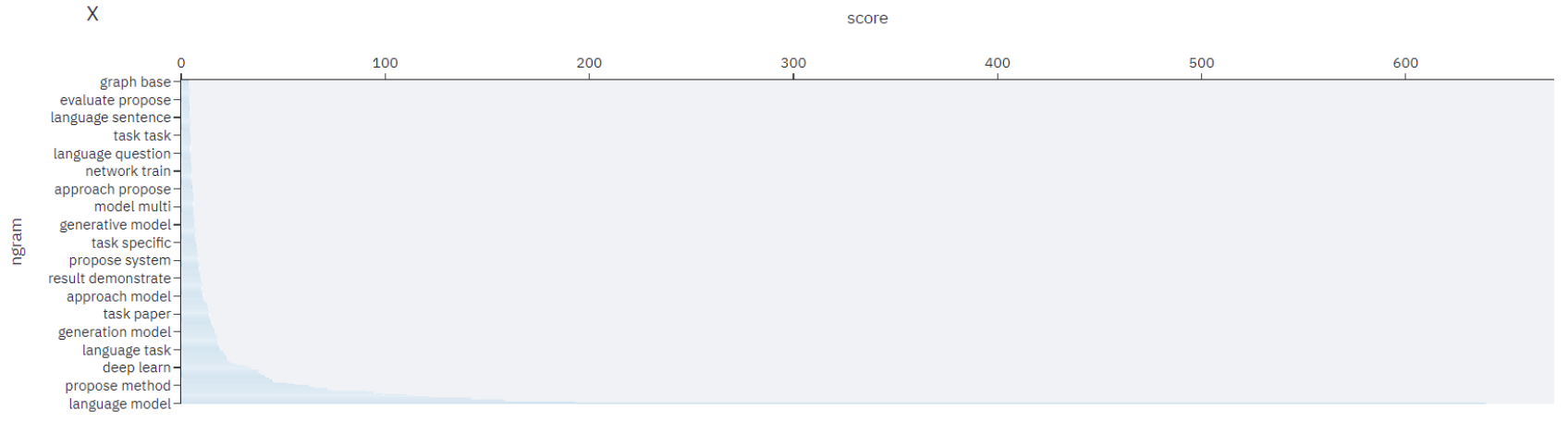}

    \caption{Keyword Visualisation using the YAKE keyword extraction method for the year 2019. Top keywords with $n=2$ include ``language model'', ``deep learn'', ``generative model'', etc.}
    \label{fig:keyword_viz}
\end{figure*}

\end{document}